\documentclass{article}

\usepackage{ms}
\usepackage[utf8]{inputenc} 
\usepackage[T1]{fontenc}    
\usepackage{hyperref}       
\usepackage{url}            
\usepackage{booktabs}       
\usepackage{amsfonts}       
\usepackage{nicefrac}       
\usepackage{microtype}      
\usepackage{lipsum}
\usepackage{microtype}
\usepackage{graphicx}
\usepackage{subfigure}
\usepackage{booktabs} 
\usepackage{amsmath}
\usepackage{csquotes}
\usepackage{amsfonts}
\usepackage{hyperref}

\usepackage{amsmath}

\DeclareMathOperator{\EX}{\mathbb{E}}
\DeclareMathOperator{\KL}{\mathbb{K}\mathbb{L}}
\newcommand\mydots{\hbox to 1em{.\hss.\hss.}}

\title{MHVAE: a Human-Inspired Deep Hierarchical Generative Model\\ for Multimodal Representation Learning}

\author{
  Miguel Vasco\\
  INESC-ID \& Instituto Superior\\ T\'{e}cnico, Universidade de Lisboa\\
  \texttt{miguel.vasco@gaips.inesc-id.pt} \\
   \And
  Francisco S. Melo\\
  INESC-ID \& Instituto Superior\\ T\'{e}cnico, Universidade de Lisboa\\
   \And
  Ana Paiva\\
  INESC-ID \& Instituto Superior\\ T\'{e}cnico, Universidade de Lisboa\\
}

\begin{document}
\maketitle

\begin{abstract}
Humans are able to create rich representations of their external reality. Their internal representations allow for cross-modality inference, where available perceptions can induce the perceptual experience of missing input modalities. In this paper, we contribute the Multimodal Hierarchical Variational Auto-encoder (MHVAE), a hierarchical multimodal generative model for representation learning. Inspired by human cognitive models, the MHVAE is able to learn modality-specific distributions, of an arbitrary number of modalities, and a joint-modality distribution, responsible for cross-modality inference. We formally derive the model's evidence lower bound and propose a novel methodology to approximate the joint-modality posterior based on modality-specific representation dropout. We evaluate the MHVAE on standard multimodal datasets. Our model performs on par with other state-of-the-art generative models regarding joint-modality reconstruction from arbitrary input modalities and cross-modality inference.
\end{abstract}


\section{Introduction}

\begin{figure}[t]
  \centering
  \includegraphics[width=8.2cm]{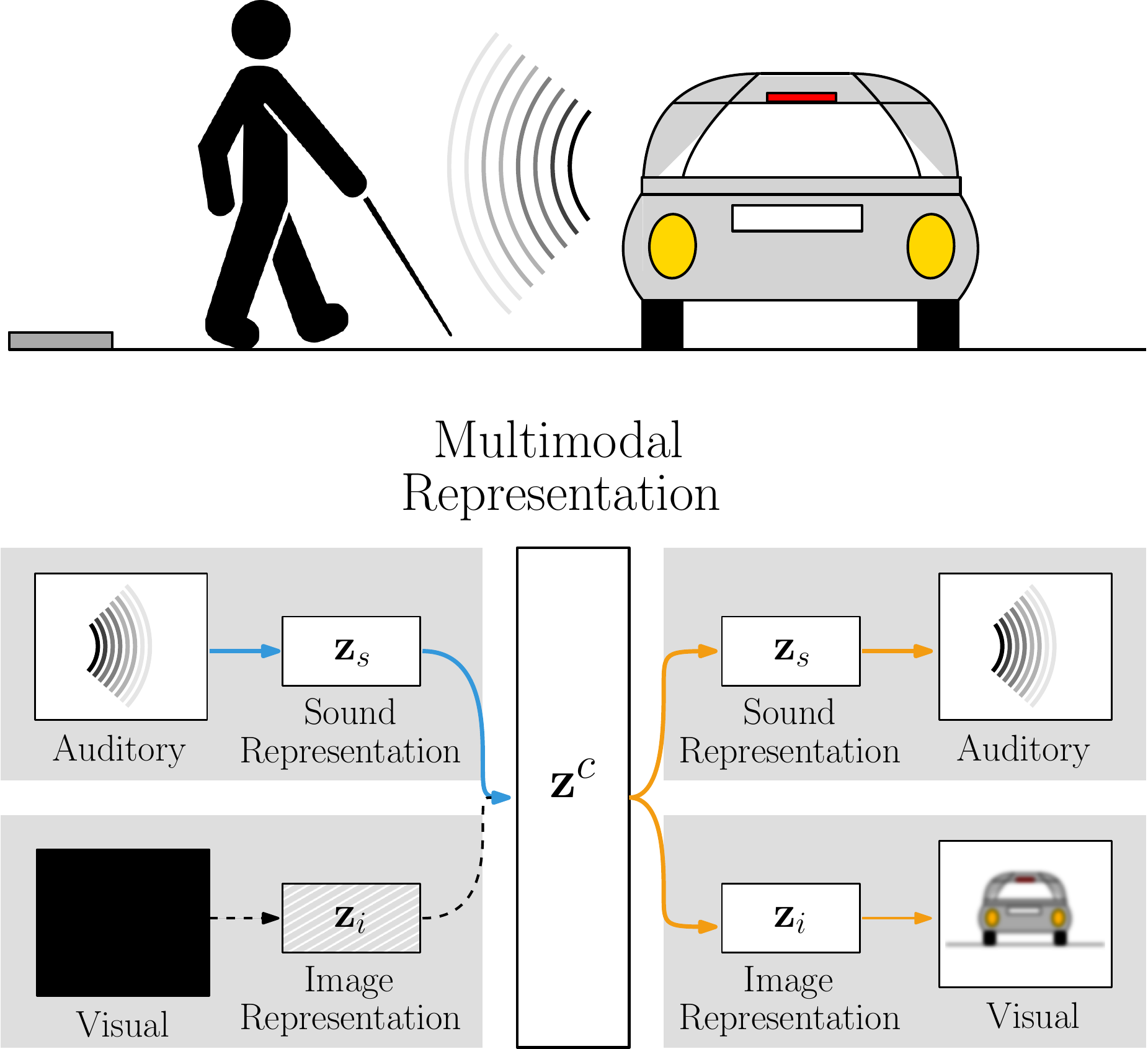}
  \caption{An example of the importance of multimodal representation learning for human tasks: in the absence of light, humans can navigate their environment by employing perceptual information from other modalities (such as sound) to generate the absent visual perceptual experience. Following human cognitive models, in this work, we contribute with the MHVAE, a novel multimodal hierarchical variational autoencoder able to perform cross-modality inference.}
  \label{fig:intro}
\end{figure}

Humans are provided with a remarkable cognitive framework which allows them to create a rich representation of their external reality. This framework contains the tools to learn novel representations of their environment and to recognize previously learned representations, which are stored in memory~\cite{meyer2009convergence, damasio1989time}. The information provided by the environment is of a multimodal nature, captured and processed by the different sensory input channels (\textit{senses}) humans possess. Yet, information is often incomplete, be it due to some modality not being provided by the environment or due to human sensory malfunction. To overcome such events, the human cognitive framework also allows for cross-modality inference, a process in which an available input modality can induce perceptual experiences of the missing modalities~\cite{walker2010preverbal,maurer2006shape,spence2011crossmodal}. Figure~\ref{fig:intro} illustrates how cross-modality inference is essential in humans to act upon their environment in scenarios of incomplete perceptual observations.

Artificial agents, on the other hand, struggle to obtain rich representations of their environment. For example, in spite of being endowed with multiple sensors, robots often disregard the multimodal nature of environmental information and learn internal representations from a single perceptual modality, often vision~\cite{finn2017deep, pinto2016curious}. However, such disregard leads to the agent's inability to understand and act upon its environment when that modality-specific information is unavailable or in the (frequent) case of sensory malfunction. If we aim at having artificial agents---such as service robots or autonomous vehicles---acting reliably in their environments, they must be provided with mechanisms to overcome potential perceptual issues. Rich joint-modality representations can play a fundamental role in robust policy transfer across different input modalities of artificial agents~\cite{silva_playing_2019}. 

\begin{figure*}
\centering
\subfigure[]{\label{Fig:CDZ_model}\includegraphics[height=60mm]{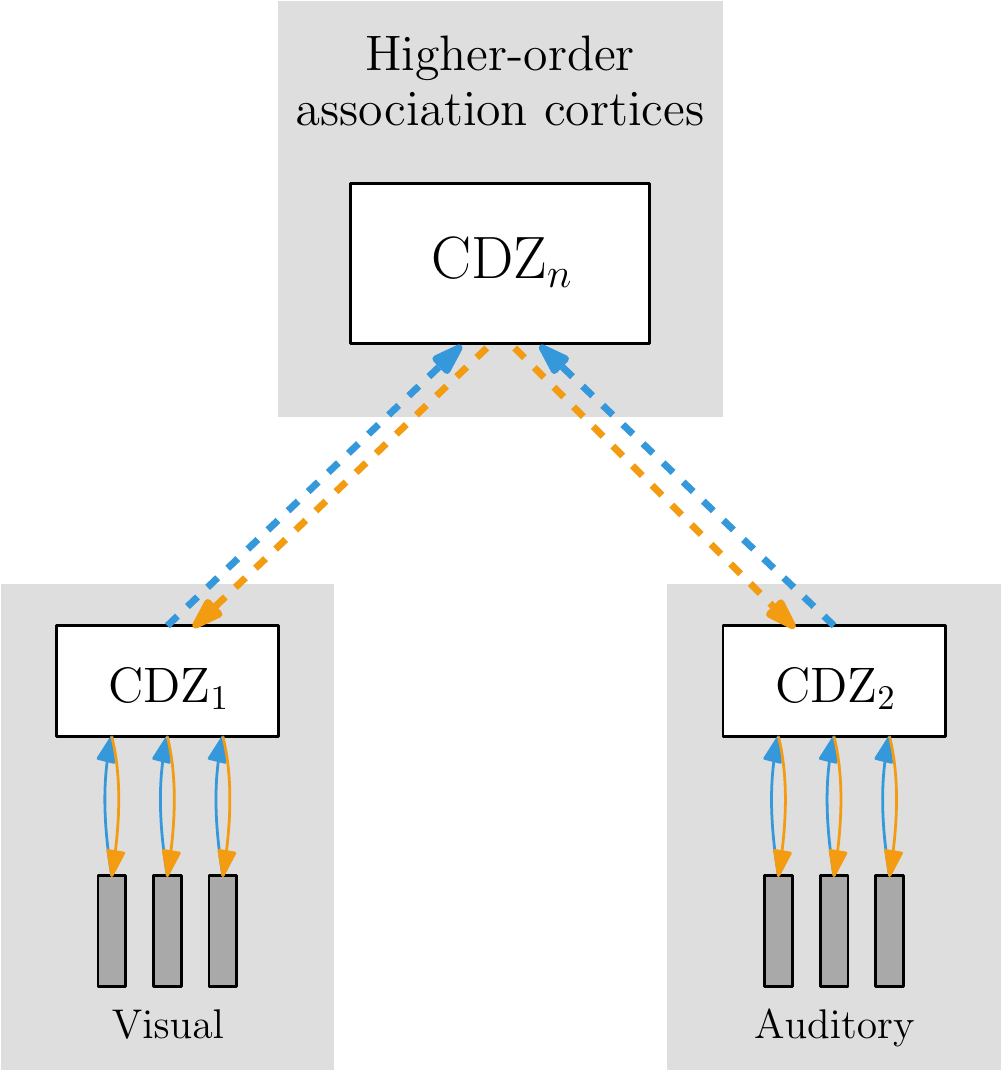}}
\quad
\quad
\quad
\quad
\subfigure[]{\label{Fig:MHVAE_model}\includegraphics[height=60mm]{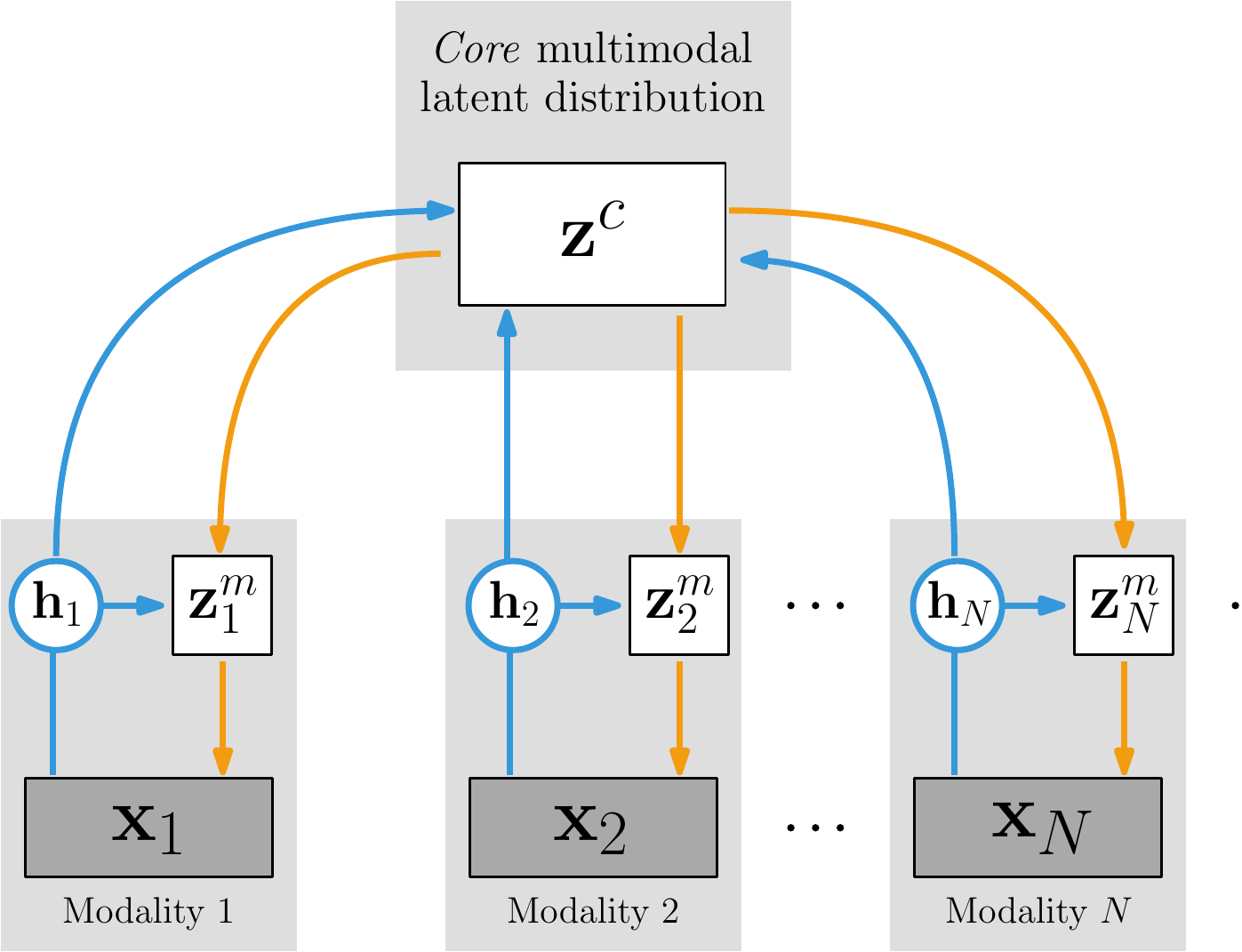}}
\caption{Schematic of the CDZ and MHVAE models, where grey rectangles indicate observed variables: (a) the inference and generative models of the CDZ model; (b) the MHVAE model, in which a high-level \textit{core} latent variable $\mathbf{z}_c$ generates the \textit{modality-specific} latent distributions $\mathbf{Z}_m = \{\mathbf{z}_1 \ldots \mathbf{z}_N\}$, responsible for sampling modality data $\mathbf{X} = \{\mathbf{x}_1 \ldots \mathbf{x}_N\}$. This generative process is represented by the orange segments. The blue segments represent the inference model of the MHVAE, where modality observations $\mathbf{X} = \{\mathbf{x}_1 \ldots \mathbf{x}_N\}$ are encoded both in the \textit{modality-specific} latent distributions $\mathbf{Z}_m = \{\mathbf{z}_1 \ldots \mathbf{z}_N\}$ and in the \textit{core} latent distribution $\mathbf{z}_c$, considering the hidden modality-specific representations $\mathbf{h} = \{\mathbf{h}_1 \ldots \mathbf{h}_N\}$ .}
\label{Fig:MHVAE_Model}
\end{figure*}

Inspired by the human cognitive framework, we contribute a novel model capable of learning rich multimodal representations and performing cross-modality inference. Multimodal generative models have shown great promise in doing so by learning a single joint-distribution of multiple modalities~\cite{shi2019variational,suzuki2016joint,korthals2019multi,wu2018multimodal}. This single representation space has to encode information to account for the complete generation process of all modalities, often of different complexities. As such, for each input modality, the representation capability of this single joint-representation space must pale in comparison with that of an individual modality-specific space. Indeed, according to the Convergence-Divergence Zone (CDZ) cognitive model~\cite{damasio1989time}, humans process perceptual information not in a single representation space but in a hierarchical structure: sensory data is processed at lower-levels of the model, generating modality-specific representations; and divergent information from these representations is merged at higher levels of the model, generating multimodal representations~\cite{meyer2009convergence, lallee2013multi}. The architecture of the CDZ model is presented in Figure~\ref{Fig:CDZ_model}.

Inspired by CDZ architecture, we propose the MHVAE, a novel generative model that learns multimodal representations in an unsupervised way. The MHVAE model is a multimodal hierarchical Variational Autoencoder (VAE) that learns modality-specific distributions, of an arbitrary number of modalities, and a joint-modality distribution, allowing for cross-modality inference. Moreover, we formally derive the model's evidence lower bound (\textsc{ELBO}) and, based on modality-specific representation dropout, we propose a novel methodology to approximate the joint-modality posterior. This approach allows the encoding of information from an arbitrary number of modalities and naturally promotes cross-modality inference during the model's training, with minimal computational cost.

We evaluate the potential of the MHVAE as a multimodal generative model on standard multimodal datasets. We show that the MHVAE outperforms other \mbox{state-of-the-art} multimodal generative models on modality-specific reconstruction and cross-modality inference.

In summary, the main contributions of this paper are:
\begin{itemize}
    \item We propose a novel multimodal hierarchical VAE, inspired by the CDZ-based human neural architecture~\cite{damasio1989time}. The model learns modality-specific distributions and a joint distribution of all modalities, allowing for cross-modality inference in the presence of incomplete perceptual information. We formally derive the model's evidence lower bound.
    
    \item We propose a new methodology for approximating the joint-modality posterior, based on modality-specific representation dropout. This approach allows the encoding of information from an arbitrary number of modalities and naturally promotes cross-modality inference during the model's training, with minimal computational cost.
    
    \item We evaluate the model on standard multimodal datasets and show that the MHVAE performs on par to other \mbox{state-of-the-art} multimodal generative models on modality-specific reconstruction and cross-modality inference.
\end{itemize}

\section{A Deep Hierarchical Generative Model for Multimodal Representation Learning}


Deep generative models have shown great promise in learning generalized representations of data. For single-modality data, the VAE is widely used. It learns a joint distribution $p_{\theta}(\mathbf{x}, \mathbf{z})$ of data $\mathbf{x}$, which is generated by a latent variable $\mathbf{z}$. This latent variable is often of lower dimensionality in comparison with the modality itself and acts as the representation vector in which data is encoded.

The joint distribution takes the form
\begin{equation*}
p_{\theta}(\mathbf{x}, \mathbf{z}) = p_{\theta}(\mathbf{x}\mid\mathbf{z}) \, p(\mathbf{z}),
\end{equation*} 
where $p(\mathbf{z})$ (the \textit{prior} distribution) is often an unitary Gaussian (\mbox{$\mathbf{z} \sim \mathcal{N}(\mathbf{0}, \mathbf{I})$}). The generative distribution $p_{\theta}(\mathbf{x}\, |\, \mathbf{z})$, parameterized by $\theta$, is usually composed with a simple likelihood term (e.g. Bernoulli or Gaussian).

The training procedure of the VAE model involves the maximization of the evidence likelihood $p(\mathbf{x})$, by marginalizing over the latent variable:
\begin{equation*}
    p(\mathbf{x}) = \int_{\mathbf{z}} p_{\theta}(\mathbf{x}, \mathbf{z}) = \int_{\mathbf{z}} p_{\theta}(\mathbf{x}\, |\, \mathbf{z}) \, p(\mathbf{z}).
\end{equation*}
However, the above likelihood is intractable. As such, we resort to an inference network $q_{\phi}(\mathbf{z}|\mathbf{x})$ for its estimation:
\begin{equation*}
    p(\mathbf{x}) = \int_{\mathbf{z}} p_{\theta}(\mathbf{x}|\mathbf{z}) p(\mathbf{z})\frac{q_\phi(\mathbf{z}|\mathbf{x})}{q_\phi(\mathbf{z}|\mathbf{x})}.
\end{equation*}
Applying the logarithm and Jensen's inequality we obtain a lower-bound on the log-likelihood of the evidence (ELBO), i.e., $\log p(\mathbf{x}) \geq \mathcal{L}(\mathbf{x})$, where
\begin{equation*}
        \mathcal{L}(\mathbf{x}) = \EX_{q_{\phi}(\mathbf{z}|\mathbf{x})} \left[ \log p_{\theta} (\mathbf{x}|\mathbf{z}) \right] - \, \KL \left[q_{\phi}(\mathbf{z} | \mathbf{x}) \, || \, p(\mathbf{z})\right],
\end{equation*}
where the Kullback-Leibler divergence term, $\KL \left[q_{\phi}(\mathbf{z} | \mathbf{x}) \,||\, p(\mathbf{z})\right]$, promotes a balance between the latent variable's capacity and the encoding process of data. During training, this balance can be adjusted through the introduction of a hyper-parameter $\beta$,
\begin{equation*}
        \mathcal{L}(\mathbf{x}) = \EX_{q_{\phi}(\mathbf{z}|\mathbf{x})} \left[ \log p_{\theta} (\mathbf{x}|\mathbf{z}) \right] - \, \beta \, \KL \left[q_{\phi}(\mathbf{z} | \mathbf{x}) \, || \, p(\mathbf{z})\right],
        \label{Eq:VAE_ELBO}
\end{equation*}
where we recover the original VAE formulation when taking $\beta=1$. The optimization of the ELBO is done using gradient-based methods, applying a re-parametrization technique~\cite{kingma2013auto}.

\subsection{MHVAE}

We now introduce the MHVAE model, which extends the single modality nature of VAEs to the multimodal hierarchical setting. In the multimodal setting, we consider a set of $N$ modalities \mbox{$\mathbf{X} = \{\mathbf{x}_1, \mathbf{x}_2, \ldots, \mathbf{x}_N\} = \mathbf{x}_{1:N}$}, generated accordingly to some environmental-dependent process $p_{\theta}(\mathbf{x}_{1:N})$, parameterized by $\theta$. We model the generation process of information in a hierarchical fashion: each modality is generated by a corresponding \textit{modality-specific} latent variable in the set $\mathbf{Z}^m = \{\mathbf{z}^{m}_{1:N}\}$, conditionally independent given a \textit{core} latent variable $\mathbf{z}^c$. The main goal of the MHVAE is to simultaneously learn single-modality latent spaces, for reconstructing modality-specific data, and a joint distribution of modalities, encoded in a core latent distribution, allowing cross-modality inference. The architecture of the proposed model is presented in Figure~\ref{Fig:MHVAE_model}.

\subsubsection{Evidence Lower-bound of the MHVAE}

In order to train the model, we aim at maximizing the likelihood of the generative process, $p_{\theta}(\mathbf{x}_{1:N})$, by marginalizing over the modality-specific and core latent variables,
\begin{equation}
  p_{\theta}(\mathbf{X}) = \int_{\mathbf{z}^c}\int_{\mathbf{z}^{m}_{1:N}} p(\mathbf{x}_{1:N}, \mathbf{z}^{m}_{1:N}, \mathbf{z}^c).
\end{equation}
Given its hierarchical nature and the conditional independence of each modality-specific latent variable in regards to the core latent variable, we can decompose the joint-modality probability as
\begin{equation}
  p_{\theta}(\mathbf{X}) = \int_{\mathbf{z}^c}\int_{\mathbf{z}^{m}_{1:N}} p(\mathbf{z}^c) \, \prod_{i=1}^{N} p_{\theta}(\mathbf{x}_i|\mathbf{z}^{m}_i)\, p_{\theta}(\mathbf{z}^{m}_i|\mathbf{z}^c).
\end{equation}
However, since the marginal likelihood of each modality is intractable, we estimate its posterior resorting to an inference model $q_{\phi}(\mathbf{z}^{m}_{1:N}, \mathbf{z}^c)$, parameterized by $\phi$. We consider an inference model \mbox{$q_{\phi}(\mathbf{z}^{m}_{1:N}, \mathbf{z}^c)$}, as shown in Figure~\ref{Fig:MHVAE_model}, in which modality information is encoded simultaneously into the modality-specific latent spaces and into the core latent space, yielding
\begin{equation}
  q_{\phi}(\mathbf{z}^{m}_{1:N}, \mathbf{z}^c) = q_{\phi}(\mathbf{z}^c | \mathbf{x}_{1:N}) \prod_{i=1}^{N} q_{\phi}(\mathbf{z}^{m}_i | \mathbf{x}_i).
\end{equation}

Introducing the inference model in the decomposed joint probability and rewriting the likelihood of the evidence as an expectation over the latent variables, we obtain
\begin{equation}
  p_{\theta}(\mathbf{x}_{1:N}) = \EX_{\substack{q_{\phi}(\mathbf{z}^{m}_{1:N} | \mathbf{x}_{1:N})\\q_{\phi}(\mathbf{z}^c | \mathbf{x}_{1:N})}}
  \left[\frac{p(\mathbf{z}^c)}{q_{\phi}(\mathbf{z}^c | \mathbf{x}_{1:N})}\right.
  \left.\prod_{i=1}^{N} \frac{p_{\theta}(\mathbf{x}_i|\mathbf{z}^{m}_i)\, p_{\theta}(\mathbf{z}^{m}_i|\mathbf{z}^c)}{q_{\phi}(\mathbf{z}^{m}_i | \mathbf{x}_i)}\right].
\end{equation}

\noindent Taking the logarithm and applying Jensen's inequality~\cite{jensen1906fonctions}, we estimate a lower-bound on the log-likelihood of the evidence $\log p_{\theta}(\mathbf{x}_{1:N}) \geq \mathcal{L}(\mathbf{x}_{1:N})$ as

\begin{equation}
  \mathcal{L}(\mathbf{X}) =  \EX_{\substack{q_{\phi}(\mathbf{z}^{m}_{1:N} | \mathbf{x}_{1:N})\\ q_{\phi}(\mathbf{z}^c | \mathbf{x}_{1:N})}}
  \log \left[\frac{p(\mathbf{z}^c)}{q_{\phi}(\mathbf{z}^c | \mathbf{x}_{1:N})}\right.
  \left.\prod_{i=1}^{N} \frac{p_{\theta}(\mathbf{x}_i|\mathbf{z}^{m}_i)\, p_{\theta}(\mathbf{z}^{m}_i|\mathbf{z}^c)}{q_{\phi}(\mathbf{z}^{m}_i | \mathbf{x}_i)}\right].
\label{Eq:MHVAE}
\end{equation}

\noindent The lower-limit $\mathcal{L}(\mathbf{x}_{1:N})$ can be seen as containing three distinct groups. The first group, similar to the original VAE formulation, corresponds to the reconstruction loss of input $\mathbf{x}_{1:N}$, generated by the modality-specific latent variables $\mathbf{z}^{m}_{1:N}$. For the $i$-th modality, this is given by
\begin{equation}
    \int\limits_{\mathbf{z}^{m}_{1:N}} \int\limits_{\mathbf{z}^c} \log p_{\theta}(\mathbf{x}_i|\mathbf{z}^{m}_i) q_{\phi}(\mathbf{z}^{m}_i | \mathbf{x}_i) q_{\phi}(\mathbf{z}^c | \mathbf{x}_{1:N}) \prod_{\substack{j=1\\j\neq i}}^{N} q_{\phi}(\mathbf{z}^{m}_j | \mathbf{x}_j)
    =\EX_{q_{\phi}(\mathbf{z}^{m}_i |\mathbf{x}_i)} \left[\log p_{\theta}(\mathbf{x}_i|\mathbf{z}^{m}_i)\right].
\end{equation}

\noindent The second component parallels the encoding capacity constrain of the latent variable in the VAE formulation, now considering the multimodal core latent variable $\mathbf{z}^c$. This constraint penalizes encoding distributions $q_{\phi}(\mathbf{z}^c \: | \: \mathbf{z}^{m}_{1:N})$ that deviate from the prior $p(\mathbf{z}^c)$ and is given by

\begin{equation}
    \int\limits_{\mathbf{z}^{m}_{1:N}} \int\limits_{\mathbf{z}^c} \log \frac{p_{\theta}(\mathbf{z}^c)}{q_{\phi}(\mathbf{z}^c | \mathbf{x}_{1:N})} q_{\phi}(\mathbf{z}^c | \mathbf{x}_{1:N}) \prod_{\substack{i=1}}^{N} q_{\phi}(\mathbf{z}_i | \mathbf{x}_i)  
    = -\KL\left[q_{\phi}(\mathbf{z}^c \: | \: \mathbf{x}_{1:N}) \,||\, p(\mathbf{z}^c)\right].
\end{equation}

\noindent Finally, the third term associates the distribution generated by the single-modality encoders, $q_{\phi}(\mathbf{z}^{m}_i|\mathbf{x}_i)$, and the distribution generated from the multimodal \textit{core} latent space, $p_{\theta}(\mathbf{z}^{m}_i | \mathbf{z}^c)$:

\begin{equation}
    \int\limits_{\mathbf{z}^{m}_{1:N}} \int\limits_{\mathbf{z}^c} \log \frac{p_{\theta}(\mathbf{z}^{m}_i|\mathbf{z}^c)}{q_{\phi}(\mathbf{z}^{m}_i | \mathbf{x}_i)} q_{\phi}(\mathbf{z}^{m}_i | \mathbf{x}_i) q_{\phi}(\mathbf{z}^c | \mathbf{x}_{1:N}) \prod_{\substack{j=1\\j\neq i}}^{N} q_{\phi}(\mathbf{z}^{m}_j | \mathbf{x}_j)
    = -\EX_{q_{\phi}(\mathbf{z}^c | \mathbf{x}_{1:N})} \big[\KL\left[q_{\phi}(\mathbf{z}^{m}_i|\mathbf{x}_i) \,||\, p_{\theta}(\mathbf{z}^{m}_i|\mathbf{z}^c)\right]\big].
\end{equation}

Taking into consideration the previous components, we can write the evidence lower-bound of the MHVAE model as
\begin{align}
\nonumber%
\mathcal{L}(\mathbf{x}_{1:N})= &\sum_{i=1}^{N} \: \lambda_i \: \EX_{q_{\phi}(\mathbf{z}^{m}_i|\mathbf{x}_i)} \left[ \log p_{\theta}(\mathbf{x}_i | \mathbf{z}^{m}_i)\right]\\ 
\nonumber%
- &\sum_{i=1}^{N} \: \beta^{m}_i \EX_{q_{\phi}(\mathbf{z}^c | \mathbf{x}_{1:N})} \big[\KL\left[q_{\phi}(\mathbf{z}^{m}_i|\mathbf{x}_i) \,||\, p_{\theta}(\mathbf{z}^{m}_i|\mathbf{z}^c)\right]\big] \\
- \: &\beta^c \: \KL\left[q_{\phi}(\mathbf{z}^c \: | \: \mathbf{x}_{1:N}) \,||\, p(\mathbf{z}^c)\right],\\
\nonumber%
\label{Eq:MHVAE_ELBO}
\end{align}
%
where we introduce weight factors $\lambda_i$ for each modality-specific reconstruction loss and a divergence term $\beta^{m}_i$, in addition to a core capacity weight $\beta^c$.

\subsubsection{Modality Representation Dropout}

\begin{figure*}
\centering
\subfigure[]{\label{Fig:MHVAE_RMD_1}\includegraphics[height=57mm]{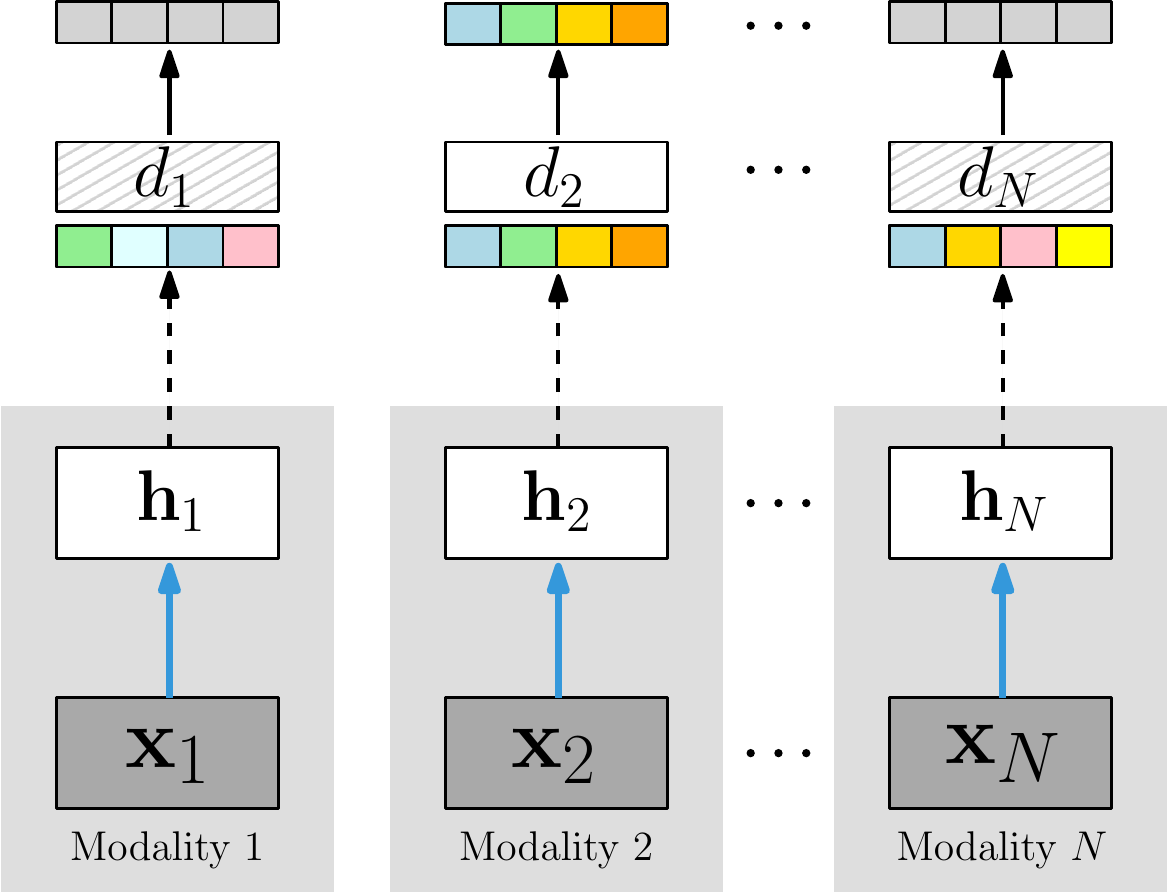}}
\quad
\quad
\subfigure[]{\label{Fig:MHVAE_RMD_2}\includegraphics[height=57mm]{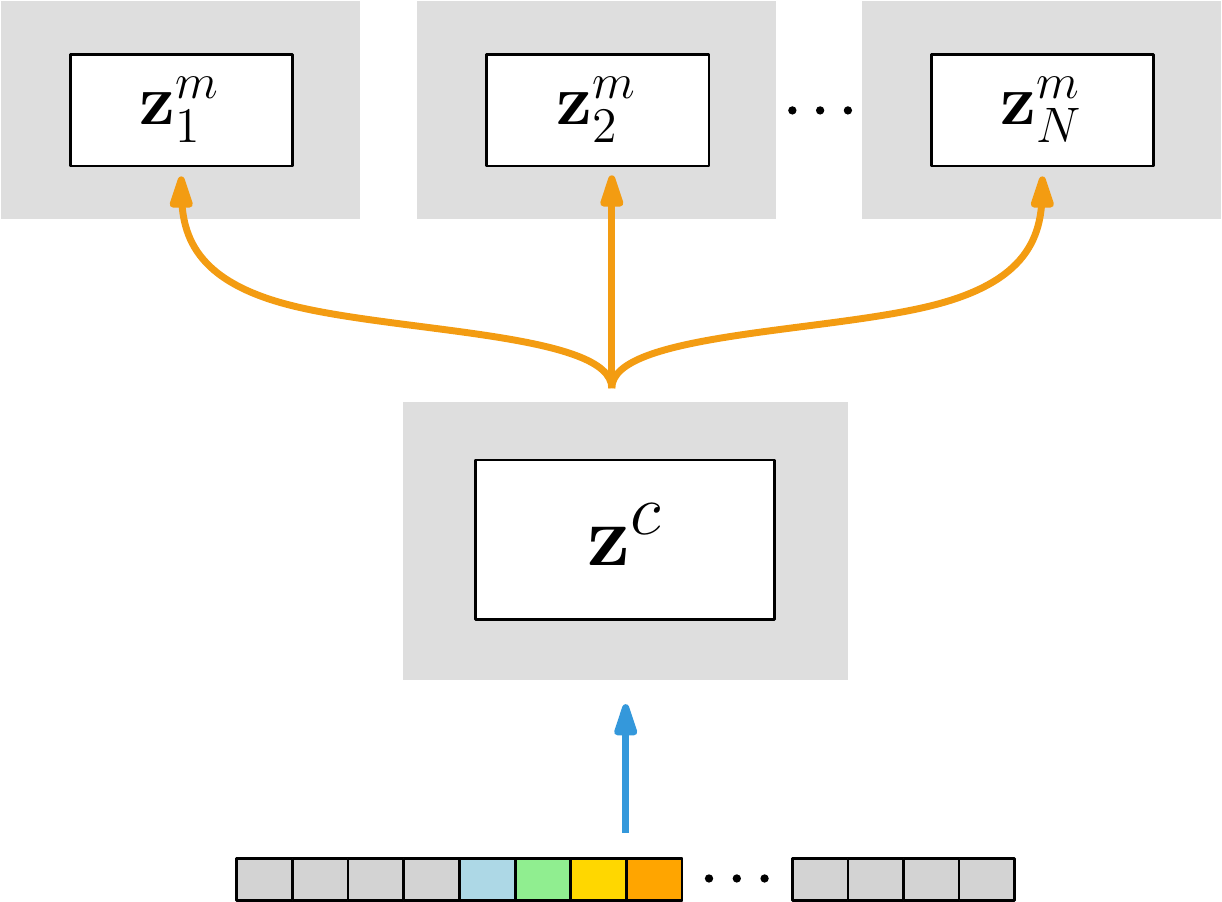}}
\caption{Diagram of the proposed Modality Representation Dropout (MRD) procedure for training the MHVAE: (a) after encoding modality observations $\mathbf{x}_{1:N}$, we compute the \textit{modality-specific} hidden representations $\mathbf{h} = \{\mathbf{h}_1, \ldots, \mathbf{h}_N\}$, sampling the dropout mask $\mathbf{d}\sim {\rm Bern}\,(\mathbf{w}_{1:N})$, in order to zero-out the selected modality representations; (b) after the procedure, we concatenate the hidden representations and encode the multimodal latent variable $\mathbf{z}^c$.}
\label{Fig:MHVAE_RMD}
\end{figure*}

We now turn to the methodology to approximate the joint-modality posterior distribution. In the case of the MHVAE, we wish to encode information from the modality-specific data $\mathbf{x}_{1:N}$ into the multimodal core latent variable $\mathbf{z}^{c}$.

One approach to do so, the product-of-experts (POE), approximates the joint posterior with the product of Gaussian experts including a prior expert\cite{wu2018multimodal}. However, this solution is computationally intensive (as it requires artificial sub-sampling of the observations during training) and suffers from overconfident expert prediction, resulting in sub-par cross-modality inference performance~\cite{shi2019variational}.

We propose a novel methodology to approximate the joint-modality posterior based on the dropout of modality-specific representations, as shown in Figure~\ref{Fig:MHVAE_RMD}. We introduce a modality-data dropout masks $\mathbf{d}$, with dimensionality \mbox{$|\mathbf{d}| = N$}, such that
\begin{equation}
\mathbf{h}^{d} = \mathbf{d} \odot \mathbf{h},
\end{equation}
where $\mathbf{h} = \{\mathbf{h}_1, \ldots, \mathbf{h}_N\}$ corresponds to the list of hidden-layer representations, computed by the modality-data encoders, as seen in Figure~\ref{Fig:MHVAE_model}. We effectively zero-out the selected components by considering that
\begin{equation}
    \mathbf{h}_i = \mathbf{0},\,\, \text{if} \,\, d_i = 1.
\end{equation}
During training, for each datapoint, we sample $\mathbf{d}$ from a Bernoulli distribution,
\begin{equation}
\mathbf{d}\sim {\rm Bern}\,(w_1, \ldots, w_N),\,\,\,\,\text{with }\sum_{i=1}^{N} d_i \geq 1,
\end{equation}
where the hyper-parameters $\mathbf{w}_{1:N}$ control the dropout probability of each modality representation. Moreover, we condition the mask sampling procedure in order to always allow at least a single modality representation to be non-zero. As such, for each sample we concatenate the resulting representations to be used as input to the multimodal encoder. Accounting for latent modality dropout, the modified ELBO of the MHVAE becomes
\begin{align}
\nonumber%
\lefteqn{\mathcal{L}(\mathbf{X})=  \, \sum_{i=1}^{N} \: \lambda_i \: \EX_{q_{\phi}(\mathbf{z}^{m}_i|\mathbf{x}_i)} \left[\log p_{\theta}(\mathbf{x}_i | \mathbf{z}^{m}_i)\right]}\\
\nonumber%
& - \sum_{i=1}^{N} \: \beta^{m}_i \EX_{q_{\phi}(\mathbf{z}^c | \mathbf{h}^{d})} \big[\KL\left[q_{\phi}(\mathbf{z}^{m}_i|\mathbf{x}_i) \,||\, p_{\theta}(\mathbf{z}^{m}_i|\mathbf{z}^c)\right]\big]\\
& - \: \beta^c \: \KL\left[q_{\phi}(\mathbf{z}^c \: | \: \mathbf{h}^{d}) \,||\, p(\mathbf{z}^c)\right].\\
\nonumber%
\label{Eq:MHVAE_ELBO}
\end{align}

\section{Evaluation}

In this section, we evaluate the MHVAE's performance as a multimodal generative model on standard multimodal datasets. Our model outperforms other state-of-the-art generative models regarding joint-modality reconstruction from arbitrary input modalities and cross-modality inference.

\subsection{Multimodal Datasets}

As in previous literature, we transform single modality datasets into bimodal datasets by considering the label associated to each image as a modality of its own right. We also compare the MHVAE to existing multimodal generative models: JMVAE-kl~\cite{suzuki2016joint} and MVAE~\cite{wu2018multimodal}. For the JMVAE-kl model we consider $\alpha = 0.01$. For the MVAE model, trained using the publicly available official implementation,%
\footnote{Implementation available at \url{https://github.com/mhw32/multimodal-vae-public}}
we employ the author's suggested training hyper-parameters.

We evaluate our model on literature standard datasets: MNIST~\cite{lecun1998gradient}, FashionMNIST~\cite{xiao2017fashion}, and CelebA~\cite{liu2018large}. We report state-of-the-art performance on the first two datasets regarding generative modelling and cross-modality capabilities.

We train the MHVAE with no hyper-parameter tuning, i.e., \mbox{$\alpha_i = \beta^{m}_i = \beta^c = 1, \forall i \in [1, N]$}. Moreover, we fix the dropout hyper-parameters $\mathbf{w}_{1:N} = 0.5$, for all modalities. For the MHVAE model, as shown in Figure~\ref{Fig:MHVAE_model}, we consider two different types of networks: the modality network, responsible for encoding the input data into the modality-specific latent space $\mathbf{z}^m$, the associated hidden representation $\boldsymbol{h}$, and the inverse generative process; and the core network, responsible for the encoding of the multimodal core latent variable, from the representation $\mathbf{h}^{d}$, from which we generate the modality-specific latent spaces $\mathbf{z}^m$. For fairness, on each dataset, we keep the network architectures consistent across
models: the generative and inference networks of the baseline models share their architecture with the modality-specific networks of the MHVAE.

Moreover, we also consider a warm-up period on the regularization terms of the ELBO~\cite{sonderby2016ladder}: we linearly increase the value of the prior regularization term on the modality-specific latent variable for $U_m$ epochs; and we linearly increase the value of the Gaussian prior on the core latent space for $U_c$ epochs. For the baselines we consider a single warm-up period on the prior regularization of the latent space, $U_b$.

We evaluate the reconstruction capabilities and cross-modality inference performance of the models. To do so, we estimate the image marginal log-likelihood, \mbox{$\log p(\mathbf{x}_1)$}, the joint log-likelihood, \mbox{$\log p(\mathbf{x}_1, \mathbf{x}_2)$}, and the conditional log-likelihood, \mbox{$\log p(\mathbf{x}_1 |\mathbf{x}_2)$}, of the observations, through importance sampling. For MNIST and FashionMNIST, we consider $5,000$ importance samples, and for CelebA we consider $500$ samples. The evaluation metrics are derived in appendix.

\subsubsection{MNIST}

\begin{table}
\centering
  \caption{Log-likelihood values of the proposed evaluation metrics on the MNSIT dataset for the MHVAE and other multimodal generative models. We estimate the latent variables considering as input image (I), label (L) or joint (I, L) modalities, resorting to $5000$ importance samples. Due to numerical instabilities we were unable to train or evaluate the MVAE baseline model.}
\begin{tabular}{@{}lcccc@{}}
\toprule
Metric                                & Input & JMVAE           & MVAE    & MHVAE             \\ \midrule
$\log p(\mathbf{x}_1)$                &I       & -90.189         & -  & \textbf{-89.050}  \\
$\log p(\mathbf{x}_1, \mathbf{x}_2)$  &I       & -90.241         & -  & \textbf{-89.183}  \\
$\log p(\mathbf{x}_1, \mathbf{x}_2)$  &L       & -125.381        & - & \textbf{-121.401} \\
$\log p(\mathbf{x}_1, \mathbf{x}_2)$  &I,L     & -90.335         & -  & \textbf{-89.143}  \\
$\log p(\mathbf{x}_1 | \mathbf{x}_2)$ &L       & -123.070        & - & \textbf{-118.856}\\\bottomrule
\end{tabular}
\label{table:mnist}
\end{table}

For the MNIST dataset, we train all models on images \mbox{$\mathbf{x}_1 \in \mathbb{R}^{28\times 28}$} and labels \mbox{$\mathbf{x}_2 \in \left\{0, 1\right\}^{10}$}. We consider a dataset division of 85\% for training, $10\%$ of which for validation purposes, and the remaining 15 $\%$ for evaluation.

We compose the image modality network of the MHVAE model with three linear layers with 512 hidden units, leaky rectifiers as activation function and applying batch-normalization between each hidden layer. Furthermore, we consider a 16-dimensional image-specific latent space. The label modality network is similarly composed with three linear layers with 128 hidden units, considering a 16-dimensional label-specific latent space.  The core network is composed with three linear layers with 64 hidden units, considering a 10-dimensional latent space. For the baselines, we consider a single 26-dimensional latent space.

\begin{figure*}
\centering
\subfigure[]{\includegraphics[height=40mm]{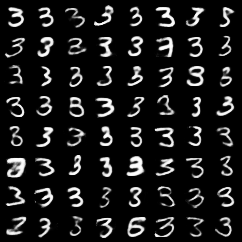}}
\quad
\quad
\quad
\subfigure[]{\includegraphics[height=40mm]{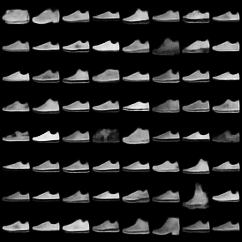}}
\quad
\quad
\quad
\subfigure[]{\includegraphics[height=40mm]{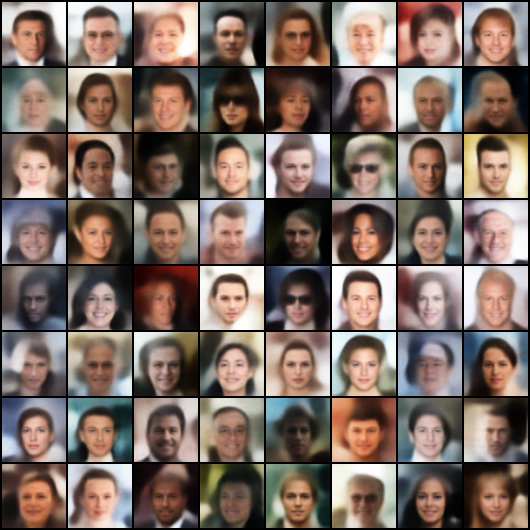}}
\quad
\quad
\quad
\subfigure[]{\includegraphics[height=40mm]{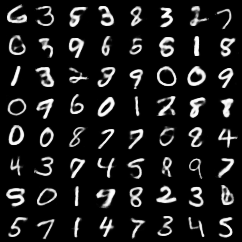}}
\quad
\quad
\quad
\subfigure[]{\includegraphics[height=40mm]{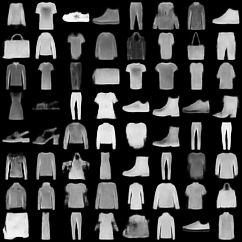}}
\quad
\quad
\quad
\subfigure[]{\includegraphics[height=40mm]{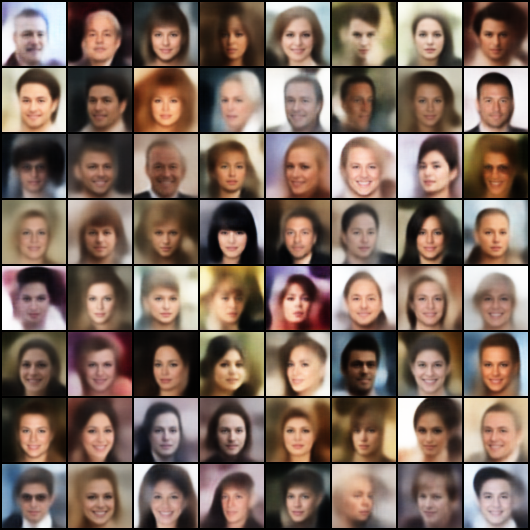}}
\caption{Image samples generated by the MHVAE model trained on standard multimodal datasets: (a), (b), (c) show images conditionally generated from sampling \mbox{$\mathbf{z}^c \sim q_{\phi}(\cdot | \mathbf{x}_2)$}, with (a) $x_2 = 3$, (b) $x_2 = $\{Sneaker\}, (c) $x_2 = $\{Male Smiling\}; and (d), (e), (f) show images generated by sampling from the prior \mbox{$\mathbf{z}_c \sim \mathcal{N}(\mathbf{0}, \mathbf{I})$.}}
\label{Fig:MuHVE_Samples}
\end{figure*}

We consider $p(\mathbf{x}_1|\mathbf{z}_1)$ as a Bernoulli distributed likelihood and $p(\mathbf{x}_2|\mathbf{z}_2)$ as a multinomial likelihood. Moreover, for the MHVAE model, we consider \mbox{$U_m = 100, U_c = 200$} epochs and for the baselines $U_b = 200$ epochs.

We train all models for 500 epochs, considering a learning rate of $l=10^{-3}$ and batch-size $b=64$. The estimates of the test log-likelihoods for all the models are presented in Table~\ref{table:mnist}. We report that the MHVAE outperforms other \mbox{state-of-the-art} multimodal models on both single-modality and joint-modality metrics, despite the fact that these separate representation spaces are of lower dimensionality than the joint representation space employed by the JMVAE and the MVAE. Moreover, the MHVAE model is able to provide better cross-modality inference than other models, as observed by the significantly lower value of the conditional log likelihood \mbox{$\log p(\mathbf{x}_1 | \mathbf{x}_2)$}, employing only the lower-dimensional label modality to estimate this quantity.

In Figure~\ref{Fig:MuHVE_Samples}, we present images generated by the MHVAE, sampled from the prior \mbox{$z_c \sim \mathcal{N}(\mathbf{0}, \mathbf{I})$} and conditioned on a given label, i.e., estimated using \mbox{$q_{\phi}(\mathbf{z}_c |\mathbf{x}_2)$}. The quality of the sampled images indicates a suitable performance of the generative networks of the MHVAE model.

\subsubsection{FashionMNIST}

\begin{table}
\centering
  \caption{Log-likelihood values of the proposed evaluation metrics on the FashionMNIST dataset for the MHVAE and other multimodal generative models. We estimate the latent variables considering as input image (I), label (L) or joint (I, L) modalities, resorting to 5000 importance samples.}
\begin{tabular}{@{}lcccc@{}}
\toprule
Metric                                & Input & JMVAE           & MVAE     & MHVAE             \\ \midrule
$\log p(\mathbf{x}_1)$                & I       & -232.427        & -236.613 & \textbf{-231.753} \\
$\log p(\mathbf{x}_1, \mathbf{x}_2)$  & I       & -232.739        & -242.628 & \textbf{-232.276} \\
$\log p(\mathbf{x}_1, \mathbf{x}_2)$  & L       & -244.378       & -557.582 & \textbf{-243.932} \\
$\log p(\mathbf{x}_1, \mathbf{x}_2)$  & I,L     & -232.573        & -241.534 & \textbf{-232.248} \\
$\log p(\mathbf{x}_1 | \mathbf{x}_2)$ & L       & -242.060        & -552.679 & \textbf{-241.662}\\\bottomrule
\end{tabular}
\label{table:fashion}
\end{table}

\noindent For FashionMNIST, we train the generative models on greyscale images \mbox{$\mathbf{x}_1 \in \mathbb{R}^{1\times 28\times 28}$}  and their class labels \mbox{$\mathbf{x}_2 \in \left\{0, 1\right\}^{10}$}, with the same proportional division of the dataset as for the previous case. For the MHVAE model, we implement a miniature DCGAN~\cite{radford2015unsupervised} architecture as the image-modality encoder, with Swish~\cite{ramachandran2017searching} as the activation function due to its performance in deep convolutional-based models. The network is composed of two convolutional layers of 32 and 64 channels followed by a linear layer of 128 hidden units. For the core and text-modality inference and generator networks, we maintain the same architecture. We consider modality-specific and core latent spaces with the same dimensionality as before and employ the same training hyper-parameters as in the previous evaluation case.

We train all models for 500 epochs, employing the Adam optimization algorithm~\cite{kingma2014adam} in the training procedure with learning rate $10^{-3}$ and batch-size $b=64$. The estimates of the test log-likelihoods for all the models are presented in Table~\ref{table:fashion}.  Once again, we report that the MHVAE outperforms other \mbox{state-of-the-art} multimodal models on both single-modality and joint-modality metrics, as well as on \mbox{label-to-image} cross-modality inference.

In Figure~\ref{Fig:MuHVE_Samples}, we present the images generated by the MHVAE, sampled from the prior \mbox{$z_c \sim \mathcal{N}(\mathbf{0}, \mathbf{I})$} and conditioned on a given label, which provide evidence that the generative networks of the model have a suitable performance.

\subsubsection{CelebA}

For CelebA, we train the MHVAE on re-scaled colored images \mbox{$\mathbf{x}_1 \in \mathbb{R}^{3 \times 64\times 64}$}  and a subset of 18 visually distinctive attributes \mbox{$\mathbf{x}_2 \in \left\{0, 1\right\}^{18}$}~\cite{perarnau2016invertible}. We compose the image modality network of the MHVAE model as a miniature DCGAN~\cite{radford2015unsupervised}. This network is composed of four convolution layers, with \mbox{${32, 64, 128,\text{ and }256}$} channels, respectively, followed by a linear layer of 512 hidden units.   Furthermore, we consider an 48-dimensional image-specific latent space. The label modality network is composed with three linear layers with 512 hidden units, considering an 48-dimensional label-specific latent space. The core network is composed with three linear layers with 256 hidden units, considering a 16-dimensional latent space. The baselines consider a single 64-dimensional latent space.

We train all models for 50 epochs employing a learning rate $10^{-4}$ and batch-size $b=128$. For the MHVAE model, we consider $U_m = 5$ and $U_c = 10$ epochs. For the baselines models, we consider a single warm-up period of $U_b = 10$ epochs. The estimates of the test log-likelihoods, computed using 500 importance samples, are presented in Table~\ref{table:celeba}. In this scenario, the MHVAE performs on par with other \mbox{state-of-the-art} multimodal models on all metrics, albeit with slight less performance in comparison with the previous evaluations. In Figure~\ref{Fig:MuHVE_Samples}, we present the images generated by the MHVAE, sampled from the prior \mbox{$z_c \sim \mathcal{N}(\mathbf{0}, \mathbf{I})$} and conditioned on a given set of attributes.

\begin{table}
\centering
  \caption{Log-likelihood values of the proposed evaluation metrics on the CelebA dataset for the MHVAE and other multimodal generative models. We estimate the latent variables considering as input image (I), attributes (A) or joint (I, A) modalities, resorting to 500 importance samples.}
\begin{tabular}{@{}lcccc@{}}
\toprule
Metric                                & Input & JMVAE           & MVAE     & MHVAE             \\ \midrule
$\log p(\mathbf{x}_1)$                & I       & -6260.35       & \textbf{-6256.65} & -6271.35 \\
$\log p(\mathbf{x}_1, \mathbf{x}_2)$  & I       & \textbf{-6264.59}        & -6270.86 & -6278.19 \\
$\log p(\mathbf{x}_1, \mathbf{x}_2)$  & A       & \textbf{-7204.36}        & -7316.12 & -7303.64 \\
$\log p(\mathbf{x}_1, \mathbf{x}_2)$  & I,A     & \textbf{-6262.67}        & -6266.14 & -6276.57 \\
$\log p(\mathbf{x}_1 | \mathbf{x}_2)$ & A       & \textbf{-7191.11}        & -7309.10 & -7296.22\\\bottomrule
\end{tabular}
\label{table:celeba}
\end{table}

\subsection{Discussion}

We have evaluated the MHVAE against a baseline of \mbox{state-of-the-art} regarding performance on standard multimodal datasets. We have compared our model with the JMVAE and the MVAE models, two widely used models for multimodal representation learning.

The results, on increasingly complex datasets, attest to the importance of considering hierarchical representation spaces to model multimodal data distributions. Even considering lower-dimensional spaces to learn the modality distributions, in comparison with the single-multimodal space of the baselines, the MHVAE is able to achieve \mbox{state-of-the-art} results on the MNIST and FashionMNIST datasets, with minimal hyper-parameter tuning.

On the CelebA dataset, the MHVAE behaves on par with the other baseline models, which raises the question about the importance of the dimensionality of the representation spaces for complex scenarios. Indeed, for a fair comparison with the other baselines, we limited the MHVAE model to have lower-dimensional representations spaces which, on a complex datasets such as CelebA, result in a lower log-likelihood of the modalities. However, the MHVAE model is still capable of outperforming the MVAE model in regards to joint-modality and cross-modality inference, estimated from the label. Regarding future work, we intend to address the question of the balance between representative capacity in the core and in the modality-specific distributions.

\section{Related Work}

Deep generative models have shown great promise in learning generalized latent representations of data. The VAE model~\cite{kingma2013auto} estimates a deep generative model through variational inference methods, encoding univariate data in a single latent space, regularized by a prior distribution. The regularization distribution is often an unitary Gaussian, or a more complex posterior distribution~\cite{su2018f, rezende2015variational}. Due to the intractability of the marginal likelihood of the data, the model resorts to an inference network in the computation of the model's evidence lower-bound. This lower-bound can be estimated, for example, through importance sampling techniques~\cite{burda2015importance}.

Hierarchical generative models have also been proposed in literature to learn complex relationships between latent variables~\cite{sonderby2016ladder, zhao2017learning,bachman2016architecture,hsu2018hierarchical}. However, these models consider representations created from a single modality and, as such, are not able to provide a framework for cross-modality inference nor to represent multimodal data. On the other hand, VAE models have also been extended in order to learn joint distributions of several modalities by forcing the estimated single-modality representations to be similar, thus allowing cross-modality inference~\cite{suzuki2016joint, yin2017associate, korthals2019multi}. However, the necessity of introducing specific divergence terms in the model's evidence lower-bound for each combination of modalities hinders its application in scenarios with a large number of modalities. Another approach introduced, was the POE inference network which reduces the number of encoding networks required for multimodal encoding~\cite{wu2018multimodal}, albeit with increased computational training cost associated. In order to provide cross-modality inference capabilities, the existing models encode information from all modalities into a single, common, latent variable space. Thus, they relinquish the generative capabilities that single-modality latent representational spaces possess. In this work, we present a novel multimodal generative model, capable of learning hierarchical representation spaces.

\section{Conclusions}

In this work, by taking inspiration from the human cognitive framework, we presented the MHVAE, a novel multimodal hierarchical generative model. The MHVAE is able to learn separate modality-specific representations and a joint-modality representation, allowing for improved representation learning in comparison with the single-representation choice of other multimodal generative models. We have shown that, on standard multimodal datasets, the MHVAE is able to outperform other \mbox{state-of-the-art} multimodal generative models regarding modality-specific reconstruction and cross-modality inference.

We also proposed a novel methodology to approximate joint-modality posterior, based on modality-specific representation dropout. With minimal computational cost, this approach allows the encoding of information from an arbitrary number of modalities and naturally promotes cross-modality inference during the model’s training. We aim at exploring scenarios with larger number of modalities in the future.

Moreover, we aim to employ the MHVAE as a perceptual representation model for artificial agents and explore its application in deep multimodal reinforcement learning scenarios, when the agent has to perform cross-modality inference to perform the task. Further inspired by human cognition and perceptual learning, we also intend to explore reinforcement learning mechanisms for the construction of the multimodal representation themselves.

\bibliographystyle{unsrt}  
\bibliography{ms}  

\end{document}